\documentclass[10pt,twocolumn,letterpaper]{article}

\usepackage{cvpr}
\usepackage{times}
\usepackage{epsfig}
\usepackage{graphicx}
\usepackage{amsmath}
\usepackage{amssymb}
\usepackage{xcolor}
\usepackage{multirow}
\usepackage{subcaption}
\usepackage{tabularx}
\usepackage{ragged2e}
\newcommand{\smallsec}[1]{ \textbf{#1:}}
\usepackage[pagebackref=true,breaklinks=true,letterpaper=true,colorlinks,bookmarks=false]{hyperref}
\cvprfinalcopy
\definecolor{darkgreen}{rgb}{0,0.5,0}
\def\cvprPaperID{90}

\ifcvprfinal\fi
\begin{document}
\title{What's in a Question: Using Visual Questions as a Form of Supervision}

\author{Siddha Ganju \and
Olga Russakovsky \\ Carnegie Mellon University
\and Abhinav Gupta
}
\maketitle

\begin{abstract}
Collecting fully annotated image datasets is challenging and expensive. Many types of weak supervision have been explored: weak manual annotations, web search results, temporal continuity, ambient sound and others. We focus on one particular unexplored mode: visual questions that are asked about images. The key observation that inspires our work is that the question itself provides useful information about the image (even without the answer being available). For instance, the question ``what is the breed of the dog?'' informs the AI that the animal in the scene is a dog and that there is only one dog present. We make three contributions: (1) providing an extensive qualitative and quantitative analysis of the information contained in human visual questions, (2) proposing two simple but surprisingly effective modifications to the standard visual question answering models that allow them to make use of weak supervision in the form of unanswered questions associated with images and (3) demonstrating that a simple data augmentation strategy inspired by our insights results in a $7.1\%$ improvement on the standard VQA benchmark.
\end{abstract}

\section{Introduction}
Supervised learning has shown great promise in developing visual AI. However, collecting manually annotated visual datasets is both challenging and expensive~\cite{ILSVRC,PASCALIJCV,COCO}. Using cheaper and weaker supervision is a growing research direction~\cite{Xu15,weinzaepfel2016towards,NEIL,LEVAN,Prest12,Urtasun14,owens16,GIS_Assisted_ECCV14}. As AI is increasingly integrated into our daily lives, computer vision systems will have access to increasingly diverse sources of information by constantly observing human-human, human-object, human-environment and human-AI interactions. Efficiently utilizing all this information will become increasingly critical as we aim to develop large-scale, accurate and adaptable visual systems.

\begin{figure}
\includegraphics[width=\linewidth]{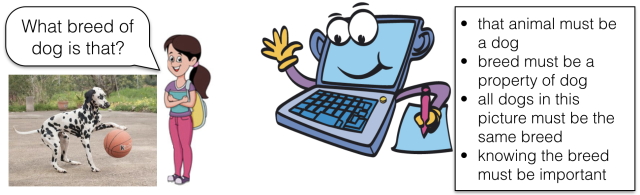}
\caption{We examine how much information is contained in a visual question, and  demonstrate that this information can be effectively used in training computer vision models.}
\label{fig:pullfig}
\end{figure}

Visual Question Answering (VQA) has become a new mode of interaction between humans and AI~\cite{VQA}. Presently, VQA is mostly used as means for evaluating visual reasoning capabilities of computers. However, going forward this is likely to become a natural human-AI interaction paradigm. It will become commonplace for humans to ask computers visual questions, such as, ``Where did I leave my keys?'', ``What breed of dog is this?'', ``Have I met this person before?'' or ``Why is she doing that?''. Instead of viewing this as single-sided interaction with humans soliciting information from AI systems, we consider how visual questions themselves can serve as a form of supervision to improve computer vision systems (Figure~\ref{fig:pullfig}).

In contrast to existing works that focus on improving AI's VQA capabilities~\cite{Fukui-park, kim-lee, Noh-han}, we strive to understand how much information is contained within the \emph{question} itself, even when the answer is not provided (as would be the case of human-AI conversations). E.g., the question ``What breed of dog is this?'' provides information that the animal in the scene is a dog and suggests that there is a single dog present. The question ``Why is he doing that?'' suggests that the depicted behavior is  unusual or unexpected. This type of free, natural and open-ended supervision can pave the way to developing richer cognitive  AI. 

We set out to investigate this hypothesis that human questions can be effectively used to improve computer vision capabilities. We begin by providing extensive qualitative and quantitative analysis of the information contained in a visual question using the large-scale VQA dataset~\cite{VQA}. We propose two simple but surprisingly effective modifications to the iBOWIMG~\cite{bow-bolei} VQA model that allows it to make use of weak supervision in the form of images associated with \emph{unanswered} questions. This proves our hypothesis that unanswered questions can be effectively used as a form of visual supervision. 

Inspired by the insights from our initial experiments, we then propose a simple data augmentation strategy. The key idea is that instead of using the image-question-answer triplet as a training exemplar, we generate $2^n$ training exemplars incorporating all possible subsets of the $n$ questions associated with the image. This strategy yields a $7.1\%$ improvement in accuracy on the standard VQA benchmark, which confirms that our analysis has important implications not just in the future of close AI-human interactions but for the immediately relevant benchmarks.

Our code, models and additional details are available at \url{http://sidgan.me/whats\_in\_a\_question/}. 

\section{Related work}
\smallsec{Vision and language} Computer vision models are generally trained to recognize a fixed vocabulary of visual concepts~\cite{PASCALIJCV,COCO,ILSVRC}. But recently, there has been a trend towards more descriptive open-world image understanding. Efforts have included works on image~\cite{Vinyals14,fang15,learning_rvr_imgcap,Vendrov16,densecap,neuraltalk2} and video~\cite{lrcn2014} captioning, image segmentation from natural language expressions~\cite{nlpsegm16}, aligning videos and books~\cite{ZhuICCV15}, zero-shot recognition from natural text~\cite{BaICCV15,Andreas-Rohrbach}, learning object models from noisy open-world human labels~\cite{Misra15} and other methods. While visual questions are certainly not meant to provide a complete description of the image, they still contain some open-world information about the scene encoded in natural text. In this work we take the initial steps towards extracting and harnessing this information. 

\smallsec{Visual question answering} The literature on building visual question answering systems~\cite{chen-wang,yang-he,xu-saenko,jiang-wang,Andreas-Rohrbach,Kafle_2016_CVPR,lu-yang,Andreas-Rohrbach-2,Shih-singh,Fukui-park,Noh-han,Ilievski-yan,wu-wang,Xiong-merity, bow-bolei,Saito-shin,kim-lee} is far too extensive to be covered in detail here. We do some analysis by building off of the iBOWIMG model of Zhou et al.~\cite{bow-bolei}. But, much of our investigation is orthogonal to the visual question answering pipeline. Our work on understanding the information embedded within a question is more similar to works such as Lin et al.~\cite{Lin_2016_ECCV} on utilizing VQA knowledge  to improve image captioning or Goyal et al.~\cite{GoyalMPB16} on analyzing the relative informativeness of different words within a question. However, in contrast to these approaches, we focus on the knowledge that can be extracted from the question alone and not the question-answer combination.

\smallsec{Incidental supervision} As we move towards large-scale open-world visual understanding, collecting manually annotated datasets for every task and concept is quickly becoming infeasible. Developing ways of using natural and cost-effective forms of supervision is a growing research direction: weak manual annotations~\cite{Xu15,weinzaepfel2016towards}, web search-based supervision~\cite{NEIL,LEVAN}, or extra modalities like temporal continuity~\cite{Prest12}, depth~\cite{Urtasun14}, ambient sound~\cite{owens16} or GPS signal~\cite{GIS_Assisted_ECCV14}. Along similar lines, we investigate whether visual questions associated with images provide sufficient supervision to train computer vision models. We argue that increasing integration of AI into everyday human environments will organically generate a large set of image-question pairs that can be used to improve visual AI systems.

\section{Inspiration: What information do unanswered questions contain?}
\label{information_content_of_question}
We begin with qualitative and quantitative analysis of the information visual questions may contain. We consider a setting where we have an image and a question (or a set of questions) associated with it, but with no corresponding answer. We examine the information content of the questions from two perspectives: (1) whether these questions can provide a good image description and (2) whether we can learn what objects are present in the image, given these questions. The insights from our analysis provide inspiration for the method described in Section~\ref{sec:method} for utilizing the unanswered questions in learning vision models.

\smallsec{Setup} We detail the setup for analysis here. We use the COCO dataset with 82,783 training and 40,504 validation images~\cite{COCO}. Three types of annotations are associated with the dataset: (1) \emph{visual questions}, where each image is associated with three human-generated questions about the visual scene~\cite{VQA}, (2) \emph{image captions}, where every image is associated with five human-generated natural language descriptions and (3) \emph{image classification labels}, where every image is annotated with the presence or absence of 80 target object classes. In this section we don't make use of the answers to the visual questions. 

\subsection{Image description}
\label{image_description}
Image captions are a natural open-world way to describe an image.~\cite{VQA} qualitatively notes the difference in information between image captions and visual questions: questions tend to provide specific information regarding one object within the image, while captions naturally tend to be a richer source of information. However, when a human looks at a scene, it is rare that she will be compelled to provide a caption (except when posting the image on social media). In contrast, she may feel compelled to ask a question, such as, ``Is this rice noodle soup?'' or ``Are the flowers real or artificial?''. The fact that she asks these questions provides some information about the scene contents. We begin by analyzing whether the visual questions contain enough information to provide an accurate description of the image.

\begin{table}[t]
\centering
\begin{tabular}{|l|l|c|c|}
\hline
Information & Model &  METEOR & SPICE  \\
\hline
\multirow{3}{*}{Qs-only} & One Q &  0.089 & 0.058 \\
& Three Qs & 0.140 & 0.115\\ 
& Seq2Seq & 0.206  &  0.140\\
\hline
Image-only & NT~\cite{neuraltalk2} & 0.267 & 0.194 \\
\hline
Image+Qs & NT + Seq2Seq &  0.305 & 0.256 \\
\hline 
\end{tabular}
\caption{Quantitative evaluation of using visual questions to provide a caption for the image, evaluated using the METEOR~\cite{meteor} and SPICE~\cite{spice2016} metrics on the  COCO validation set for the image captioning task. Details in Section~\ref{image_description}.}
\label{table:img-captioning-results}
\end{table}

\smallsec{Quantitative results} We evaluate using visual questions as image captions in Table~\ref{table:img-captioning-results} using two standard captioning metrics: METEOR~\cite{meteor} and SPICE~\cite{spice2016}. We first consider three baselines that don't use image information but generate a caption purely based on the visual questions that have been asked: (1) \emph{One Q}: using one of the visual questions directly as a caption, (2) \emph{Three Qs}: using all three visual questions concatenated together as a caption and (3) \emph{Seq2Seq}~\cite{seq2seq}: a model trained on the COCO training set that takes an input of three visual questions and learns to output an image caption based on the information contained in the questions. \emph{Three Qs} outperforms \emph{One Q} (SPICE score of 0.115 vs 0.058), indicating that different questions provide complementary information about the image content.\footnote{The SPICE metric~\cite{spice2016} considers both precision and recall of a caption, enabling a fair comparison between captions of different lengths.}
Training the \emph{Seq2Seq} model to generate more semantically meaningful captions from the three questions provides an improvement to the SPICE score from 0.115 to 0.140.~\footnote{Our analysis bears some similarity to the work of Lin et al.~\cite{Lin_2016_ECCV} on re-ranking image captions using VQA; however, we don't use the \emph{answers} or the image and evaluate captions generated solely based on the questions.}

We additionally investigate whether visual questions can provide complementary information to what is contained in the image features. We use a computer vision model NeuralTalk2 (\emph{NT})~\cite{neuraltalk2} that takes in an image and outputs an image caption. Directly concatenating this image-based caption with the caption generated from questions (\emph{NT + Seq2Seq}) improves the SPICE score from 0.194 to 0.256, indicating that the signal from visual questions may be complementary to the information in the image. 

\begin{figure}
\begin{tabularx}{\linewidth}{c@{\extracolsep{\fill}}c@{\extracolsep{\fill}}c}
\includegraphics[height=50pt]{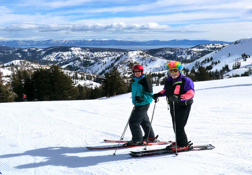} &
\includegraphics[height=50pt]{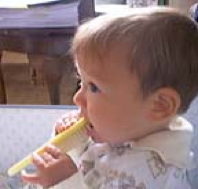} &
\includegraphics[height=50pt]{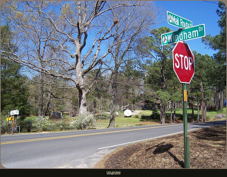} \\
\begin{minipage}[t]{0.31\linewidth}
\scriptsize
What are these two people doing in the scene? What color is the person on the right's hat? Was this picture taken during the day? \\
{\color{green} people during day with hat} 
\end{minipage} &
\begin{minipage}[t]{0.25\linewidth}
\scriptsize
What is the baby chewing on? Is this child under 5? Is the child asleep? \\
{\color{green} chewing baby}
\end{minipage} &
\begin{minipage}[t]{0.29\linewidth}
\scriptsize
Is the street name on top an unusual name for a street ? Is there a box for newspaper delivery? What color is the road? \\
{\color{green} street}
\end{minipage} 
\\
\includegraphics[height=50pt]{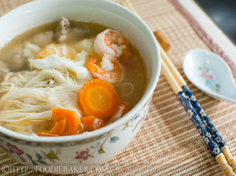} &
\includegraphics[height=50pt]{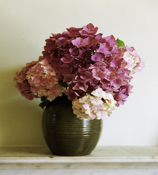} &
\includegraphics[height=50pt]{pics/Picture5.png} \\
\begin{minipage}[t]{0.31\linewidth}
\scriptsize
Is this rice noodle soup? What is to the right of the soup? What website copyrighted the picture? \\ {\color{orange} copyrighted noodle soup }
\end{minipage}  &
\begin{minipage}[t]{0.27\linewidth}
\scriptsize
Are the flowers in a vase? How many different color flowers are there? Are the flowers real or artificial? \\ {\color{orange} many green flowers}
\end{minipage}  &
\begin{minipage}[t]{0.29\linewidth}
\scriptsize
Is this a 3-D photo? How many lights on the lamppost? Is this building an unusual color? \\ {\color{orange} lampost color is light}
\end{minipage} 
\end{tabularx}
\caption{Three visual questions and the generated captions using the \emph{Seq2Seq} model in Section~\ref{image_description}. Some captions are surprisingly accurate (green) while others, less so (orange). }
\label{fig:seq2seq}
\end{figure}

\smallsec{Qualitative results} Finally, we qualitatively show some results of captions generated from visual questions. Figure~\ref{fig:seq2seq} shows some results of applying the Seq2Seq model on the validation set to convert the 3 visual questions associated with the image to a single image caption. The results demonstrate that the visual questions can provide detailed information about the image content. The generated captions contain object category, human actions, color and affordance information indicating that this information can be readily extracted from the questions. 

\subsection{Object Classification}
\label{object_classification}

Besides image description, another source of information that the visual questions can provide is the object classes that are present in the image. Some examples are shown in Table~\ref{table:coco-question-category}: e.g., asking ``what color is the bus?'' indicates the presence of a bus in the image.

\smallsec{Algorithm} To quantify how often this occurs, we extract object labels for the 80 COCO classes from visual questions. There are 64 question types in COCO: ``how many,'' ``is there'', ``what color is'', etc. For each type, we manually determine if questions of this type imply the presence of objects. For example,  questions of the type ``how many'' do imply the presence of objects:  ``how many different flowers are on the table?'' implies the presence of flowers and a table. In contrast, questions of the type ``is there,'' such as ``is there a zebra in the photo?'', do not imply the presence of any object. For each question that implies the presence of the object, we extract which of the 80 COCO classes (if any) the question refers to. We use NLTK~\cite{nltk} to disambiguate tenses and synonyms as well as \texttt{pattern.en}\footnote{\url{http://www.clips.ua.ac.be/pages/pattern-en}} for singular-plurals. For two-word categories such as ``teddy bear'' we use n-gram overlap. More details in Section \ref{algorithm_extracting_obj_from_questions}

\begin{table}[t]
\centering
\begin{tabular}{|lc|}
\hline
Question & Object class \\
\hline
What color is the {\bf bus}? & \includegraphics[height=.12in]{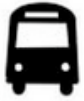} \\
Are {\bf people} waiting for the food {\bf truck}? & \includegraphics[height=.12in]{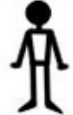} \includegraphics[height=.12in]{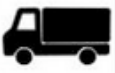}\\
How many {\bf umbrellas} are in the image? & \includegraphics[height=.12in]{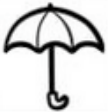} \\ 
Is the {\bf bird} sitting on a {\bf plant}? & \includegraphics[height=.12in]{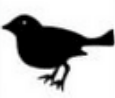} \includegraphics[height=.12in]{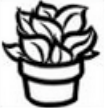} \\
\hline
\end{tabular}
    \caption{Examples of visual questions that indicate the presence of certain objects in the image.} 
    \label{table:coco-question-category}
\end{table}

\begin{figure}[t]
\centering
\includegraphics[width=0.93\linewidth]{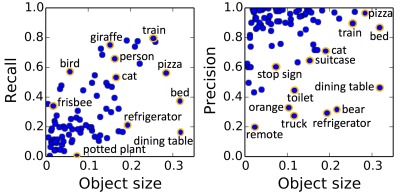}
\caption{We use the questions associated with the image to determine the objects that the image contains. We show the per-class recall \emph{(left) }  and precision \emph{(right) } of this method on the 80 COCO classes. The x axis corresponds to the average size of this object in the image. (If multiple instances of the same object class appear, we sum their areas to compute the total area occupied by this class per image.) We observe that larger objects are asked about more frequently in the questions and thus have higher recall. Most classes have $>80\%$ precision, with a few notable exceptions such as ``remote'' which can refer to the target object as well as serve as an adjective, thus having low precision of $19.3\%$. }
\label{fig:area-recall}
\end{figure}

\smallsec{Results} We compare the resulting object class vectors with ground truth annotations of object presence in the image. Our conversion algorithm achieves mean per-class recall of $29.3\%$ and precision of $82.4\%$, indicating that while the three visual questions do not refer to \emph{all} objects in the image, they nevertheless capture more than a quarter of the common objects with a few false positives. 

Figure~\ref{fig:area-recall} shows the per-class recall and precision as a function of the average size of this class in an image. As expected, objects which are larger tend to be asked about more frequently. For example, ``baseball glove'' occupies only $0.8\%$ of the image on average and has a near-zero recall of $0.7\%$, indicating that it is never asked about (or we are not able to parse it out with our algorithm). In contrast, ``train'' occupies $25.8\%$ of the image area and has a recall of $79.5\%$, indicating that if a train appears in the image it is almost always asked about. A notable exception is ``dining table,'' which occupies $31.7\%$ of the image area on average but has a recall of only $16.3\%$ since it is rarely a target object of interest. Overall across all classes, the objects we do detect occupy $18.2\%$ of the image area on average, whereas the objects that we fail to detect occupy only $7.1\%$.

\smallsec{Combination with vision models} We additionally verify that knowing the questions provides extra information about the image beyond what can currently be extracted by modern computer vision models. We finetune an ILSVRC2012-pretrained GoogLeNet model~\cite{Szegedy14,ILSVRC} on the training set of COCO to recognize the 80 target object classes. It achieves image classification mAP of $53.1\%$ on the validation set. We then combine the 80-dimensional classifier prediction vector $x_c$ with our object class vector $x_o$ extracted from the three visual questions using $\max(x_o,x_c)$. This significantly increases image classification accuracy to $67.2\%$ mAP.

\begin{figure}
\begin{tabularx}{\linewidth}{c@{\extracolsep{\fill}}c@{\extracolsep{\fill}}c}
\includegraphics[height=50pt]{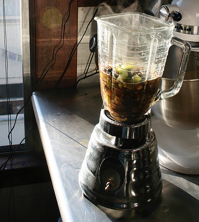} &
\includegraphics[height=50pt]{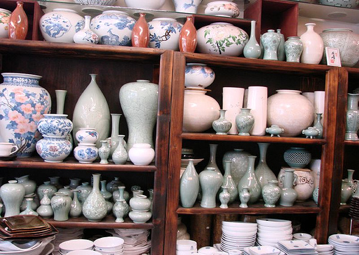} &
\includegraphics[height=50pt]{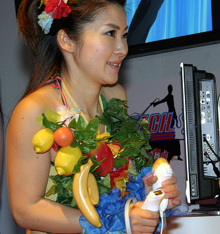} \\
\begin{minipage}{0.28\linewidth}
\scriptsize
{\color{blue} Common Sense:} 
Do the long shadows suggest it is well past morning? 
\end{minipage}  &
\begin{minipage}{0.31\linewidth}
\scriptsize
{ \color{blue} Ambiguity: }  How many identical pinkish-tan vases are on the top shelf?
\end{minipage} &
\begin{minipage}{0.28\linewidth}
\scriptsize
{\color{blue} Composition: } Is the woman's costume made of real fruit and leaves? 
\end{minipage} 
\\
\includegraphics[height=50pt]{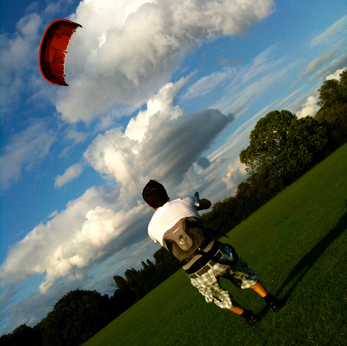} &
\includegraphics[height=50pt]{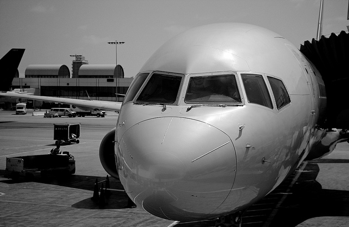} &
\includegraphics[height=50pt]{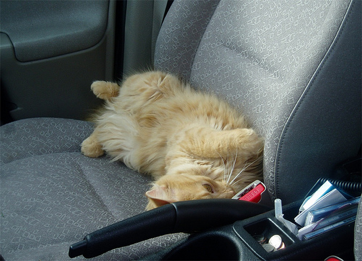} \\
\begin{minipage}{0.29\linewidth}
\scriptsize
{\color{blue} Visual Relationship:} Is the person falling down?
\end{minipage} &
\begin{minipage}{0.3\linewidth}
\scriptsize
 {\color{blue} History:} Was this picture taken over 100 years ago? 
\end{minipage}  &
\begin{minipage}{0.25\linewidth}
\scriptsize
{ \color{blue} Affordances:} Is this cat lying on the sofa?
\end{minipage} \end{tabularx}
\caption{The fact that a human was prompted to ask the question suggests that there is a relationship between the question and the image. The blue text is the type of latent information and the black is an example question.}
\label{fig:latentinfo}
\end{figure}

\smallsec{Discussion} We showed that visual questions, even without answers, provide informative image descriptions and object classification information. In addition, we briefly note that visual questions can also provide additional latent information as illustrated in Figure~\ref{fig:latentinfo}. We make no attempt to quantify here but note that this information may also potentially be extracted and exploited in the future AI systems. 

\section{Method: Effectively utilizing information from unanswered visual questions}
\label{sec:method}

Armed with the conclusion that visual questions themselves provide important and useful information about the image content, we now set out to investigate how these questions can be used to aid the development of improved computer vision models. We focus on the VQA task and investigate how even \emph{unanswered} questions can be effectively utilized to improve VQA capabilities. Since our proposed formulation is very simple, the empirical benefits demonstrated in Section~\ref{sec:experiments} are even more striking.

Standard VQA systems~\cite{chen-wang, yang-he, xu-saenko, jiang-wang, Andreas-Rohrbach, Kafle_2016_CVPR, lu-yang, Andreas-Rohrbach-2, Shih-singh, Fukui-park, Noh-han, Ilievski-yan, wu-wang, Xiong-merity, bow-bolei, Saito-shin} take the image and its target question as input, with the expectation of producing an accurate answer for the question. In Section~\ref{information_content_of_question} we made two key observations: (1) different visual questions provide information complementary to each other and (2) visual questions can provide information about the scene that may be complementary to what can be extracted from the image using modern computer vision models. Thus it is natural to ask -- can we build a better question answering system that benefits from having access to not only the image information and the target question, but also to a set of other questions that may have been asked about this image. 

\subsection{Model}
\label{2bow}
To investigate, we build upon the iBOWIMG model~\cite{bow-bolei}. This model is perfect for our investigation as it is very simple to modify and analyze, while achieving impressive results on the VQA task. iBOWIMG models the image using deep features extracted from an ILSVRC-pretrained CNN~\cite{Szegedy14,ILSVRC} and the target question using a one-hot bag-of-words text feature which is transformed via a word embedding layer. The image and target question features are concatenated and sent through a softmax layer to predict the answer class amongst a set of choices. 

\begin{figure}[t]
    \includegraphics[width=\linewidth]{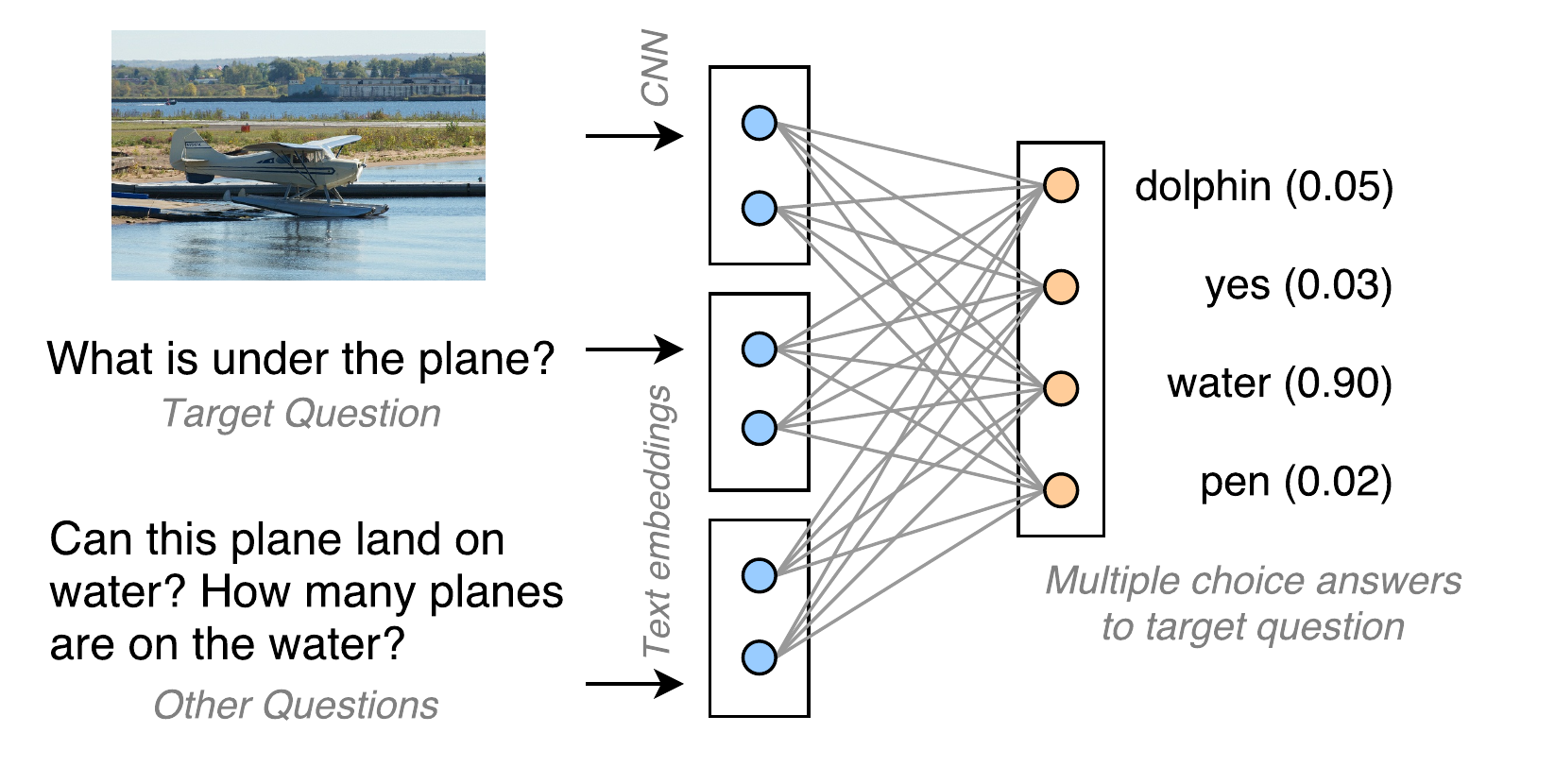}
    \caption{Framework of the iBOWIMG-2x model. The representation consists of three parts: (1) visual image features, (2)  text embedding of the target question, and (3) text embedding of the other questions concatenated together. This representation is passed through a learned fully connected layer to predict the answer to the target question.}
    \label{fig:ibowimg2x}
\end{figure}

We extend iBOWIMG to additionally take other questions which are asked about this image as input. We model these extra questions the same way as the target question: the extra questions are concatenated together into a long string, a bag-of-words text feature is computed and then transformed via a word embedding layer. This additional feature vector is concatenated with the image and target question features, as in Figure~\ref{fig:ibowimg2x}. We refer to this model as \textbf{iBOWIMG-2x} due to the increased dimensionality.\footnote{Our model bears some similarity to that of \cite{JabriJM16} which explores a different setting, where they double the dimensionality of the bag-of-words textual representation. However, they concatenate the question, image and answer features to predict the correctness of such image-question-answer triplets. In contrast, our feature vector utilizes the image features, target question and other questions about the image.}

During training the model is tasked with predicting the answer to the target question and the other questions can be thought of as a richer feature representation of the image.

\subsection{Training}
\label{sec:training}
To train the richer iBOWIMG-2x model, we need to generate new training exemplars out of the available training data. Concretely, every image $x_i$ comes associated with a set of questions $\{q_{ij}\}_j$ and corresponding answers $\{a_{ij}\}_j$. In addition, the image can also be associated with unanswered questions $\{q'_{ik}\}_k$. Let $Q^{all}_i = \{q_{ij}\}_j \cup \{q'_{ik}\}_k$ be the set of all questions associated with an image. 

The training examples for iBOWIMG are of the form:
\begin{equation}
(x_i, q_{ij}, a_{ij})\;  \forall \, i, \, j
\end{equation}

In contrast, the training examples for iBOWIMG-2x are:
\begin{equation}
(x_i, q_{ij}, E, a_{ij})\;  \forall \,i, \, j, \, E \subseteq \mathcal{P}(Q^{all}_i)
\label{eq:powerset}
\end{equation}
where $\mathcal{P}$ denotes the powerset of $Q^{all}_i$ and defines the extra information provided to the model in the form of additional questions asked about the same image.

For example, consider an image $x$ with a question $q$, a corresponding answer $a$ and two additional unanswered questions $q'_{1}$ and $q'_{2}$. For iBOWIMG, the single training example corresponding to this image would be $(x, q, a)$. For iBOWIMG-2x there would be eight training examples, with  $E$ = $\emptyset$, $q$, $q'_{1}$, $q'_{2}$,  $[q, q'_{1}]$,  $[q, q'_{2}]$,  $[q'_{1}, q'_{2}]$ or $[q, q'_{1}, q'_{2}]$ making use of the extra information that is available about this image during training in the form of other asked questions.\footnote{While this model is formulated to make use of extra unanswered questions, an additional benefit is that it can be considered a form of data augmentation. For example, if 3 answered questions are available for this image, the iBOWIMG would have 3 training examples while iBOWIMG-2x would have 24 training examples.} The target label for all these exemplars is the answer $a_{ij}$. After the new exemplars are generated, the model is trained using stochastic gradient descent exactly as iBOWIMG.

\subsection{Unanswered questions on novel images}
\label{sec:finetuning}
One disadvantage of the method described so far is that it can only incorporate information from extra questions on images that have at least one answered question provided. However, it may be the case that we have access to a large collection of images with only unanswered questions associated with them: e.g. A dataset of image-question pairs without their associated ground truth can naturally emerge from a deployed VQA system that is interacting with users in the real world. Motivated by the findings of Section~\ref{object_classification}, we use $Q^{all}_i$ to learn an image representation that may be better suited for the VQA task. Instead of using an ILSVRC-trained visual model, we use a visual model trained to recognize the words that appear within the questions. Intuitively, ILSVRC-trained models may not reflect the full spectrum of visual concepts or diverse visual scenes. This new image model can be incorporated into iBOWIMG-2x (or even iBOWIMG) as a better image representation.

\subsection{Testing} 
The iBOWIMG-2x model can be evaluated in one of two ways. During test time for a standard VQA formulation, the model only has access to a novel image $x$ and a single target question $q$. In this case, we can simply pass a zero-initialized vector for the extra features, reducing iBOWIMG-2x back to iBOWIMG but trained differently. However, iBOWIMG-2x allows additional flexibility by utilizing unanswered questions even at test time. For example, when the test image is provided with several target questions, they can further help interpret the image: e.g., test questions ``Who is to the left of the dog?'' and ``What is to the right of the person?'' provide complementary information that might help answer both questions better.

\section{Experiments}
\label{sec:experiments}
We now empirically verify our intuition that even unanswered questions can significantly improve the accuracy of VQA systems. In particular, we evaluate our proposed iBOWIMG-2x model trained on subsets of the COCO~\cite{COCO} dataset corresponding to two different settings: (1) where every image has at least one answered question and optional unanswered questions associated with it in Section~\ref{sec:exp1}, and (2) where some images have only unanswered questions associated with it in Section~\ref{sec:exp2}. We convincingly demonstrate that including extra questions significantly improves VQA accuracy. To conclude, we apply our insights to the standard VQA benchmark in Section~\ref{sec:exp3}.

\smallsec{Setup} We use the COCO dataset with 82,783 training and 40,504 validation images. Each image is associated with three questions and their corresponding answers, although we sometimes use only a subset of those in our experiments (details below). We evaluate the model on the multiple choice VQA task. We normalize the visual features and the two textual features independently to have $L2$ norm of 1. We build upon the code released by~\cite{bow-bolei}.

\subsection{Unanswered questions on training images}
\label{sec:exp1}
\smallsec{Dataset} Consider the setting where we have access to a set of training images, each with one answered question and optional unanswered questions. We simulate this by using the VQA dataset, where each training image $x_i$ is associated with 3 questions $q_{i1}, q_{i2}, q_{i3}$ and their respective answers $a_{i1},a_{i2},a_{i3}$. We randomly select a single question per image to be the target question and discard the other answers, leaving $N$ training images $x_i$, each with a question $q_i$, an answer $a_i$ and two additional unanswered question $q'_{i1}$ and $q'_{i2}$. We train the model on the COCO training image and evaluate on the validation set. Here we use GoogLeNet~\cite{Szegedy14} trained on ILSVRC2012~\cite{ILSVRC} as the visual representation.  
  
\begin{figure}
\begin{tabularx}{\linewidth}{c@{\extracolsep{\fill}}c@{\extracolsep{\fill}}c}
\includegraphics[height=50pt]{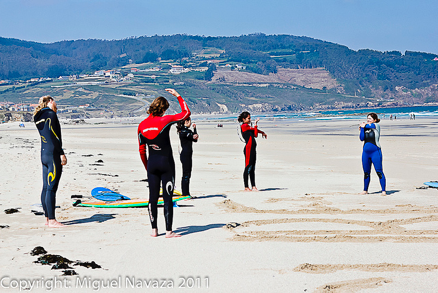} &
\includegraphics[height=50pt]{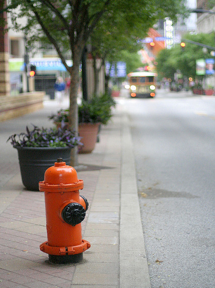} &
\includegraphics[height=50pt]{pics/Picture35.png} \\
\begin{minipage}[t]{0.35\linewidth}
\scriptsize \centering
Are these people exercising?\\
 {\color{green} Yes  }{ \color{green}   Yes }
\end{minipage} &
\begin{minipage}[t]{0.31\linewidth}
\scriptsize \centering
What object is in focus? \\
{\color{green} Fire Hydrant } {\color{red} 3}
\end{minipage} &
\begin{minipage}[t]{0.35\linewidth}
\scriptsize \centering
How many dolls are there? \\
{\color{red} No }{ \color{green} 2 }
\end{minipage} 
\\
\includegraphics[height=50pt]{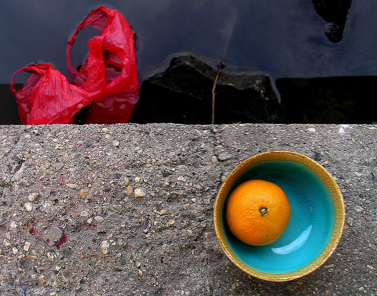} &
\includegraphics[height=50pt]{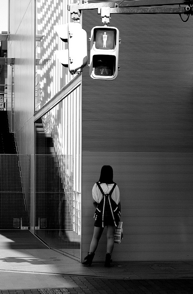} &
\includegraphics[height=50pt]{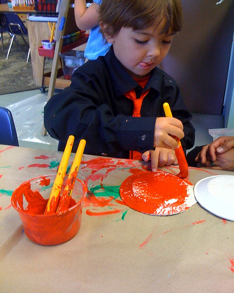} \\
\begin{minipage}[t]{0.35\linewidth}
\scriptsize \centering
What is in the water? \\
{\color{green} Plastic Bag} {\color{red} Fish } 
\end{minipage}  &
\begin{minipage}[t]{0.35\linewidth}
\scriptsize \centering
What's the girl facing?\\
{\color{green} Wall }{ \color{green} Wall}
\end{minipage}  &
\begin{minipage}[t]{0.35\linewidth}
\scriptsize \centering
Is he being messy?\\
{ \color{green} Yes }{\color{red} Red }
\end{minipage}
\\
\includegraphics[height=50pt]{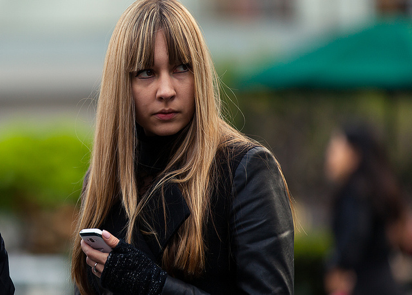} &
\includegraphics[height=50pt]{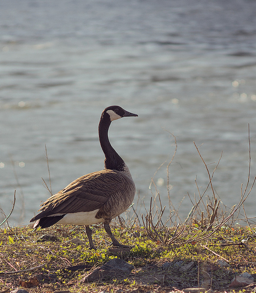} &
\includegraphics[height=50pt]{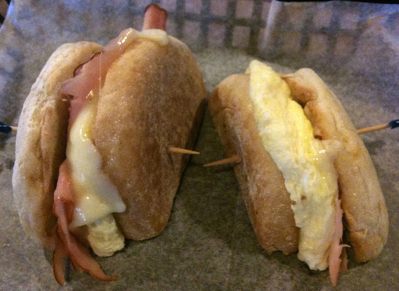} \\
\begin{minipage}[t]{0.31\linewidth}
\scriptsize \centering
What is the woman looking at so seriously? \\
{\color{red} Woman  }{ \color{green}    Person } 
\end{minipage}  &
\begin{minipage}[t]{0.31\linewidth}
\scriptsize \centering
What type of goose is pictured? \\
{\color{green}  Canadian } {\color{red} Red}
\end{minipage}  &
\begin{minipage}[t]{0.31\linewidth}
\scriptsize \centering
What is stuck in the sandwich?\\
{\color{red} No }{ \color{green} Toothpicks}
\end{minipage} 
\\
\includegraphics[height=50pt]{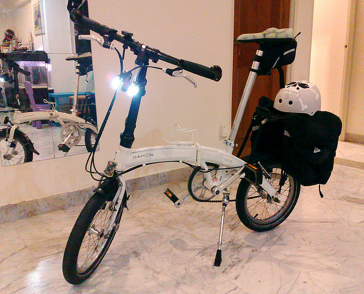} &
\includegraphics[height=50pt]{pics/Picture33.png} &
\includegraphics[height=50pt]{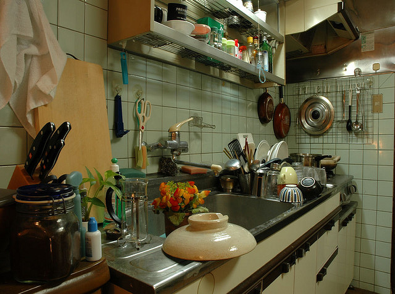} \\
\begin{minipage}[t]{0.31\linewidth}
\scriptsize \centering
 What is on the back  of the bike? \\
{ \color{green} Helmet }   { \color{red} Life Vests }
\end{minipage}  &
\begin{minipage}[t]{0.31\linewidth}
\scriptsize \centering
What color the shower curtain? \\
{\color{green} White } { \color{red} 2}
\end{minipage}  &
\begin{minipage}[t]{0.31\linewidth}
\scriptsize \centering
How many knives are in the knife holder? \\
{\color{red} 3 }{ \color{red} 6}
\end{minipage}
\end{tabularx}
\caption{Qualitative comparison of our iBOWIMG-2x \emph{(left)} and the baseline iBOWIMG \emph{(right)}. Correct answers in green; wrong answers in red. Details in Section~\ref{sec:exp1}. }
\label{fig:qual}
\end{figure}

\smallsec{Key experiment} We begin by comparing our iBOWIMG-2x model trained with the extra unanswered questions against the iBOWIMG model of~\cite{bow-bolei} which does not use the available unanswered question. After training our dataset with one answered question per image, iBOWIMG obtains an accuracy of $47.3\%$ on the validation set.\footnote{Here we evaluate the model in the standard setting where at test-time only the one target question is provided and the model is expected to produce an answer; we do this by inputting a zero-initialized vector as the second textual feature in the model (in place of the unanswered questions). } In contrast, our model makes effective use of the provided unanswered questions and achieves a significant $3.1\%$ improvement, boosting accuracy to $50.4\%$. We use bootstrapping to establish statistical significance.~\cite{PASCALIJCV} The $0.999$ confidence interval for the baseline model is $[46.6\%, 48.2\%]$; thus the improved accuracy of $50.4\%$ when including unanswered questions is statistically significant at the $\alpha = 0.001$ level. Figure~\ref{fig:qual} demonstrates qualitative results.

\smallsec{Ablation studies} We investigate two components of our model in terms of accuracy improvement: (1) the impact of having access to extra unanswered questions at training time (2) the impact of generating extra training examples per image with data augmentation based on the powerset in Eqn.~\ref{eq:powerset}. Table~\ref{table:ablation} shows the results. 

\begin{table}[t]
\centering
\begin{tabular}{|l|cc|}
\hline
Unanswered questions & Accuracy w/o aug & Accuracy \\
\hline
None & 47.34 & 47.37 \\
1 question & 48.74 & 48.94 \\
2 questions & 49.19 &  {\bf 50.37}  \\
\hline
\end{tabular}
\caption{Accuracy of iBOWIMG-2x trained with one answered question per image and optional unanswered questions. Models are trained with and without data augmentation of Eqn.~\ref{eq:powerset}. The ``None w/o aug'' setting is equivalent to iBOWIMG~\cite{bow-bolei}.  Details in Section~\ref{sec:exp1}. 
}
\label{table:ablation}
\end{table}

First, as observed above, adding the two additional unanswered questions boosts accuracy by $3.1\%$ from $47.3\%$ to $50.4\%$. It is further encouraging to note that using just one unanswered question achieves about half the improvement: a $1.6\%$ boost from the $47.3\%$ baseline to $48.9\%$ accuracy with our model. This suggests that adding more unanswered questions (which will become freely available in real-world settings) is likely to further improve accuracy.

Second, we investigate the extent of improvement due to data augmentation. Instead of using the data augmentation strategy of Eqn.~\ref{eq:powerset}, we simply train iBOWIMG-2x with a single training exemplar $(x_i,q_i,[q'_{i1} q'_{i2}],a_i)$ per image where the two extra questions are concatenated together. This yields an accuracy of $49.2\%$, which is $1.2\%$ lower than the $50.4\%$ accuracy of the whole model.\footnote{A natural question is whether this improvement arises from seeing the $(x_i, q_i, \emptyset, a_i)$ examples during training since the model is evaluated on test examples of the form $(x, q, \emptyset)$. A model trained with augmentation except \emph{without} the $(x_i, q_i, \emptyset, a_i)$ examples achieves $50.3\%$ accuracy, indicating this effect is minor. }

This suggests that although most of the improvement comes from simply having access to the extra questions, the fact that the extra questions allow us to generate a diverse augmented training set is in itself a meaningful observation. We explore this further in Section~\ref{sec:exp3}. 

\smallsec{Analysis} Digging deeper, we seek to understand what makes iBOWIMG-2x  more effective than iBOWIMG. First, we train a \emph{text-only} model which learns to answer questions without looking at the image. In this setting, iBOWIMG-2x achieves an accuracy of $47.3\%$, which is only a marginal improvement over iBOWIMG's $46.7\%$ accuracy. This suggests that much of the benefit of iBOWIMG-2x is in learning to make better use of the image features. We investigate this further in Section~\ref{sec:exp2}. 

Second, we note that iBOWIMG-2x is more likely to predict answers corresponding to actual words as opposed to a number or yes/no. In particular, iBOWIMG predicts a word answer $72.1\%$ of the time, while iBOWIMG-2x predicts a word answer only $60.2\%$ of the time. Further, iBOWIMG-2x predicts number answers at about half the rate of iBOWIMG: $12.4\%$ compared to $23.7\%$. This suggests that our model's richer representation better correlates the image appearance with the semantic textual features, making it more likely to predict a word answer instead of resorting to a simpler numerical or yes/no response.

Table~\ref{table:answer_types_validation} documents the breakdown of accuracy by answer type. Having access to the extra supervisory signal yields a $1.0\%$ improvement on \emph{number} questions, a bigger $3.0\%$ improvement on the \emph{yes/no} questions, and a large $3.9\%$ improvement on the challenging \emph{word-response} questions. Our model is unable to use unanswered questions to learn how to count object instances much better than the baseline; however, it becomes significantly better at identifying the presence or absence of visual concepts and at answering more general visual questions.

\begin{table}[t]
\centering
\begin{tabular}{|c|c|c|c|c|}
\hline
Model & Overall &  Number & Yes/No & Word\\ \hline
iBOWIMG  &  45.87  & 26.85 & 74.53 & 34.07\\ %
iBOWIMG-2x & {\bf 50.37} &  {\bf 27.92} & {\bf 77.54} & {\bf 37.98} \\
\hline 
\end{tabular}
\caption{Accuracy for each answer type. The models are trained with one answered question per image, but the iBOWIMG-2x also makes use of 2 unanswered questions.}
\label{table:answer_types_validation}
\end{table}

\smallsec{Test-time supervision} Finally, an additional advantage of our model is that it can incorporate multiple questions at test time. Concretely, instead of asking a single test question $q$ on test image $x$ and passing in the tuple $(x,q,\emptyset)$ to the model, we consider including other test questions $q_1'$ and $q_2'$ and passing in the tuple $(x,q,[q_1', q_2'])$. This yields an additional $0.5\%$ improvement in accuracy: from $50.4\%$ accuracy (when tested the standard way with only the target question available) to $50.9\%$ accuracy (when all three questions are available simultaneously).

\begin{figure*}[t]
\centering
\begin{tabularx}{\linewidth}{c@{\extracolsep{\fill}}c@{\extracolsep{\fill}}c}
\includegraphics[width=0.25\linewidth]{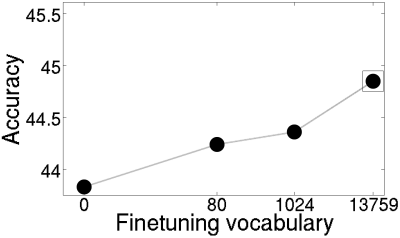} &
\includegraphics[width=0.25\linewidth]{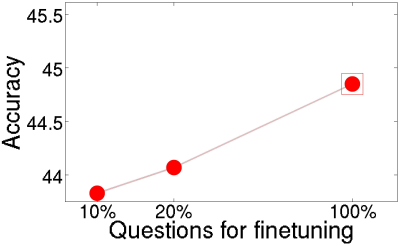} &
\includegraphics[width=0.25\linewidth]{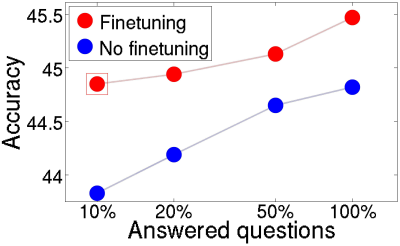}
\end{tabularx}
\caption{Effectively using training images containing only unanswered questions to improve VQA accuracy by learning the visual representation. The three squares correspond to the same model. Details in Section~\ref{sec:exp2}. 
}
\label{fig:finetuning}
\end{figure*}

\subsection{Unanswered questions on novel images}
\label{sec:exp2}

\smallsec{Dataset} In Section~\ref{sec:exp1} we considered the setting where an answered question is available on every training image. In contrast, here we consider the real-world scenario where some images have only unanswered questions associated with them. To simulate this setting, we randomly select $10\%$ of the training images to be associated with answered questions and we use only unanswered questions on the rest. We evaluate on the full validation set.

\smallsec{Key experiment} We use all the available questions to train a visual representation better suited for the VQA task. We use the pretrained AlexNet~\cite{Krizhevsky12,ILSVRC} for the baseline and compare it with the same network finetuned on the COCO training images to recognize $13,759$ words from the question vocabulary instead of the 1000 ILSVRC classes. We use these networks as the visual representation when training the iBOWIMG-2x model on the small set of available images with answered questions. The baseline network achieves $43.8\%$ accuracy; the finetuned network effectively utilizes the information captured in the unanswered questions to improve by $1.1\%$ to accuracy of $44.9\%$. 

\smallsec{Ablation studies} We evaluate two components of the framework. First, we check whether the full vocabulary is required or if filtering to 80 words (corresponding to the COCO annotated object categories and extracted from the questions as in Section~\ref{object_classification}) or 1024 words (corresponding to the most relevant words according to tf-idf extracted using the code of~\cite{bow-bolei}) would suffice. Figure~\ref{fig:finetuning}\emph{(left)} demonstrates continuous improvement with using larger vocabulary sizes. Second, we evaluate whether the full set of unanswered questions is necessary or a smaller subset would suffice. Figure~\ref{fig:finetuning}\emph{(center)} demonstrates that using more questions for finetuning progressively improves accuracy.

\smallsec{Benefits of answered vs unanswered questions} We ask one final question: how much does training a better visual representation help compared to collecting more answered questions. In Figure~\ref{fig:finetuning}\emph{(right)} we consider progressively increasing the number of available images with answered questions and compare the models with and without finetuning. Interestingly, a model finetuned with only $10\%$ of answered questions achieves an accuracy of $44.9\%$ which is on par with $44.8\%$ accuracy of the model trained on all $100\%$ answered questions without finetuning. This suggests that perhaps much of the information is already captured in the questions themselves even without the answers. However, further study is necessary to verify this claim.

\subsection{Data augmentation for VQA}
\label{sec:exp3}

Our findings demonstrate a very simple but effective way of improving VQA accuracy by adding extra unanswered questions. We take this one step further and ask the straightforward question -- can we consider the full dataset but use our model as a form of data augmentation, where all questions are used as supervisory signals at training time. Thus, we train iBOWIMG-2x where every image-question-answer triplet is now represented by 8 training exemplars. We use the setup of~\cite{bow-bolei} where the entire COCO training set and $70\%$ of the validation set is used for training. The finetuned GoogLeNet~\cite{Szegedy14} model is used for the visual representation. We evaluate on the test-dev set as standard with only one question at a time provided during testing.

iBOWIMG-2x outperforms the baseline iBOWIMG model by an impressive $7.1\%$: from $55.7\%$ for iBOWIMG to $62.8\%$ with iBOWIMG-2x having access to the exact same training question-answer pairs but with data augmentation.\footnote{Zhou et al.~\cite{bow-bolei} reports $61.7\%$ accuracy on test-dev using iBOWIMG. However, despite our best efforts, we were unable to replicate that result. Evaluating their released predictions file on test-dev obtains the same $55.7\%$ accuracy as with our retrained iBOWIMG model.}  Table~\ref{table:test-dev} documents the breakdown by answer type. The results are consistent with the findings of Section~\ref{sec:exp1}; in fact, they are even more pronounced. By making effective use of all the questions jointly through data augmentation, the model improves by $3.1\%$ on the \emph{number} questions, by $4.2\%$ on the \emph{yes/no} questions, and by an impressive $10.5\%$ on \emph{other} questions. This suggests that the data augmentation strategy may be even more beneficial for the open-ended VQA task but we leave that for future work.

These experiments demonstrate that our findings  provide important insights not only for the weakly supervised setting but also for the fully supervised VQA scenario. For completeness, our iBOWIMG-2x model achieves $63.17\%$ on test-standard. While this is not state-of-the-art accuracy, the significant $7.1\%$ improvement over the very simple model we started with suggest that our insights may be beneficial for improving the current best models as well.

\begin{table}[t]
\centering
\begin{tabular}{|c|c|c|c|c|}
\hline
Model Name  & Overall & Other & Number & Yes/No \\ \hline
iBOWIMG  &  55.68 & 42.61 & 34.87 & 76.49 \\
iBOWIMG-2x & {\bf 62.80} & {\bf 53.11} & {\bf 37.94} & {\bf 80.72} \\
\hline 
\end{tabular}
\caption{Multiple choice VQA accuracy on test-dev.}
\label{table:test-dev}
\end{table}

\section{Conclusions}
We study a previously unexplored setting of using visual questions themselves as a form of supervision to improve computer vision models. We provide both qualitative and quantitative analysis of how much information is contained within the questions. Our insights already yield significant improvements over baselines on standard benchmarks. More importantly, we believe that visual questions will become freely available as a result of human-AI interactions and can serve as a form of supervision for improving visual models. This work is an early step in this direction. 

\subsubsection*{Acknowledgements}
We would like to thank Peiyun Hu, Achal Dave, Arvind Ramachandran, Gunnar Atli Sigurdsson and Siddharth Santurkar for helpful discussions. This research was supported by ONR MURI N000141612007.

%%%%%%%%%%%%%%%%%%%%%%%%%%%%%%%%%%%%%%%%%%%%%%%%%%%%%%%%%%%%%%%%%%%%%%%%%%%%%%%%%%%%%%%%%%%%%%%%%%%%%%%%%%%%%%%%%%%%%%%%%%%%%%%%%%%%%%%%%%%%%%%%%%%%%%%%%%%%%%%%%%%%%%%%%%%%%%%%%%%%%%%%%%%%%%%%%%%%%%%%%%%%%%%%%%%%%%%%%%%%%%%%%%%%%%%%%%%%%%%%%%%%%%%%%%%%%%%%%%%%%%%%%%%%%%%%%%%%%%%%%%%%%%%%%%%%%%%%%%%%%%%%%%%%%%%%%%%%%%%%%%%%%%%%%%%%%%%%%%%%%%%%%%%%%%%%%%%%%%%%%%%%%%%%%%%%%%%%%%%%%%%%%%%%%%%%%%%%%%%%%%%%%%%%%%%%%%%%%%%%%%%%%%%%%%%%%%%%%%%%%%%%%%%%%%%%%%%%%%%%%%%%%%%%%%%%%%%%%%%%%%%%%%%%%%%%%%%%%%%%%%%%%%%%%%%%%%%%%%%%%%%%%%%%%%%%%%%%%%%%%%%%%%%%%%%%%%%%%%%%%%%%%%%%%%%%

\renewcommand\thesection{\Alph{section}}
\setcounter{section}{0}

\section*{Appendix}

\section{Extracting  Objects from the Questions}

In order to investigate the nature and quantity of information provided by the question, we use the object classification task. The Microsoft Common Objects in COntext (MS COCO) dataset contains 91 common object categories and  82 of  them  have  more  than  5,000  labeled  instances. The 2014 release considers a subset of 80 categories from  the original 91 categories. The 11 excluded categories are: hat, shoe, eyeglasses (due to too many instances in the dataset), mirror, window, door, street sign (as they were ambiguous and difficult to label), plate, desk (due to confusion with bowl and dining table, respectively), blender and hair brush (too few instances in the dataset). We strategically extract words that indicate the 80 object categories. 

\subsection{Algorithm}
\label{algorithm_extracting_obj_from_questions}

Here we provide more details for Section \ref{object_classification}. We classify the 64 question types in COCO to question types that confirm the presence of an object (e.g., ``how many different flowers are on the table?'' implies the presence of flowers and table) and those that do not (e.g., ``is there a zebra in the photo'' does not confirm the presence of any object). Figure \ref{unconfirmed_question_types} shows the question types which do not confirm the presence of an object, while Figure \ref{question_types} shows the confirmed question types. Some additional techniques to boost precision and recall which were derived from a detailed analysis of the questions are described below: \\

\begin{figure*}[h]
\begin{tabular}{cccc}
\includegraphics[height=60pt]{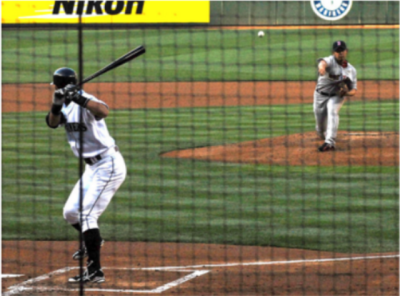} &
\includegraphics[height=60pt]{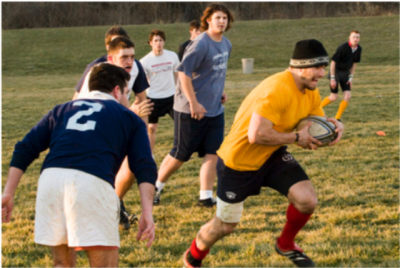} &
\includegraphics[height=60pt]{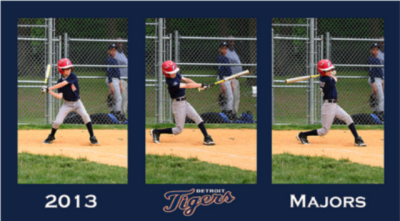} &
\includegraphics[height=60pt]{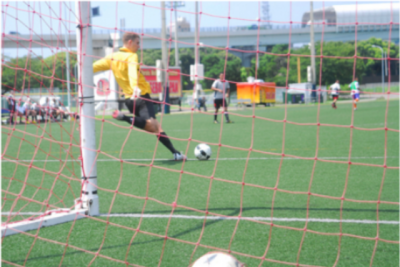} \\
\begin{minipage}[t]{0.23\linewidth}
\scriptsize \centering
Has the batter already hit the ball?
\end{minipage} &
\begin{minipage}[t]{0.23\linewidth}
\scriptsize \centering
Are the men playing rugby or football?
\end{minipage} &
\begin{minipage}[t]{0.23\linewidth}
\scriptsize \centering
Is the baseball player at home plate?
\end{minipage} &
\begin{minipage}[t]{0.23\linewidth}
\scriptsize \centering
Which foot will kick the soccer ball?
\end{minipage} \\
\includegraphics[height=60pt]{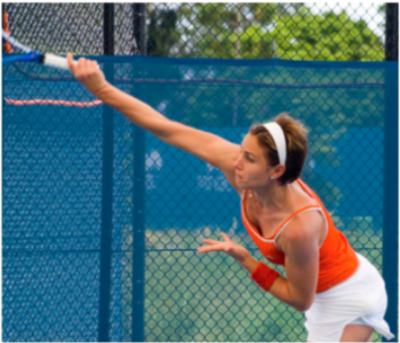} &
\includegraphics[height=60pt]{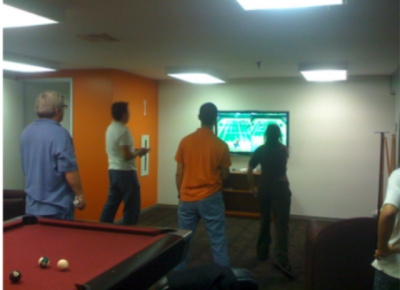} &
\includegraphics[height=60pt]{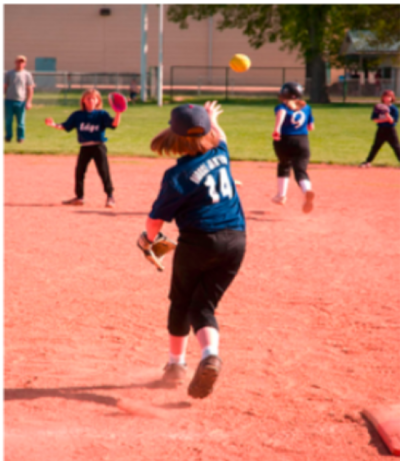} &
\includegraphics[height=60pt]{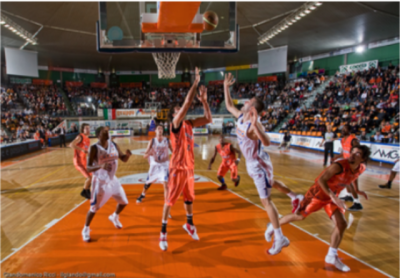} \\
\begin{minipage}[t]{0.23\linewidth}
\scriptsize \centering
Is this woman balancing herself as she hits the ball?
\end{minipage} &
\begin{minipage}[t]{0.23\linewidth}
\scriptsize \centering
How many billiard balls are there?
\end{minipage} &
\begin{minipage}[t]{0.23\linewidth}
\scriptsize \centering
What is the number on the shirt of the girl throwing the softball?
\end{minipage} &
\begin{minipage}[t]{0.23\linewidth}
\scriptsize \centering
Are the people in the stadium basketball fans?
\end{minipage} \\
\end{tabular}
\caption{True Positives for \emph{sports ball} category. This category includes various kinds of sports balls like `football', `basketball' or `baseball' and indicates that at least one \emph{sports ball} is present in the image, however the questions do not indicate any particular type of \emph{sports ball}}
\label{fig:sportsball}
\end{figure*}

\begin{figure}[h]
\begin{tabular}{c}
\includegraphics[height=60pt]{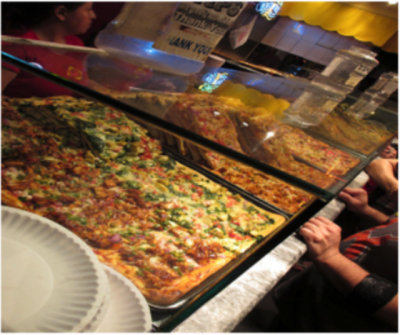} \\
\begin{minipage}[t]{\linewidth}
\scriptsize \centering
Are there enough slices of pizza to feed a football team?
\end{minipage} 
\end{tabular}
\caption{False Positives for \emph{sports ball} category.}
\label{fig:sportsball-fp}
\end{figure}

\smallsec{Super Category} The \emph{sports ball} category covers a broad spectrum as it includes all instances of various kinds of sports balls. The questions are not annotated with \emph{sports ball}, but rather with `football', `basketball' or `baseball'. For this reason, the synset category of `sports ball' is modified to have all these entities. True positives are shown in Figure \ref{fig:sportsball}, however none of the associated questions with these true positives indicate which type of \emph{sports ball} is in the image. Figure \ref{fig:sportsball-fp} shows the false positive of the \emph{sports ball} category. Similarly, the \emph{airplane} category has annotations of `jet plane', `plane', `passenger plane' and `private plane'. These are included in the synsets of `airplane'.  \\

\smallsec{Spell Check} Many words in the English language have different spellings. Due to this, category names like \emph{hair drier} are be linked to questions that contain the string `hair dryer'. Different words can be used to convey the same object or event depending on the context. For instance, the category \emph{traffic light} should cover questions that contain `traffic signal' and `traffic light'. The synsets for `traffic light' are modified to include `traffic signal'. \\ 

\begin{figure*}[h]
\begin{tabular}{cccc}
\includegraphics[height=60pt]{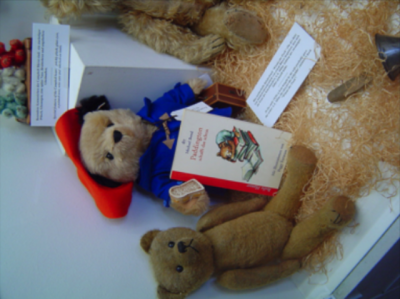} &
\includegraphics[height=60pt]{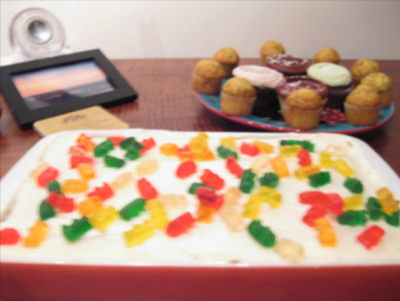} &
\includegraphics[height=60pt]{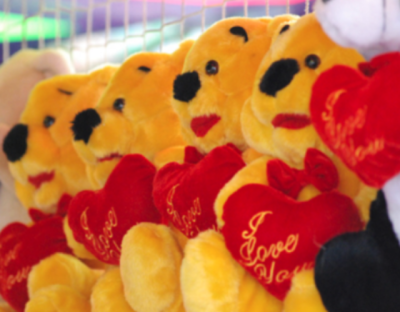} &
\includegraphics[height=60pt]{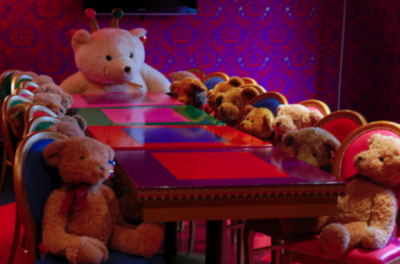} \\
\begin{minipage}[t]{0.23\linewidth}
\scriptsize \centering
Are the bears vertical or horizontal?
\end{minipage} &
\begin{minipage}[t]{0.23\linewidth}
\scriptsize \centering
What is the percentage of yellow gummy bears?
\end{minipage} &
\begin{minipage}[t]{0.23\linewidth}
\scriptsize \centering
Does the bear love you?
\end{minipage} &
\begin{minipage}[t]{0.23\linewidth}
\scriptsize \centering
Is the large bear a chairman?
\end{minipage} \\
\end{tabular}
\caption{False Positive for \emph{bear} category. The current detection algorithm does not take these cases into account. Higher precision for \emph{bear} can be obtained by carefully checking all cases. }
\label{fig:bears_false_positives}
\end{figure*}

\smallsec{Phrase-word categories}
In order to separate overlap between single word and double word categories like, \emph{bear}-\emph{teddy bear}, \emph{dog}-\emph{hot dog}, we limit the possible signals from each detected word. While detecting these, we ensure that only one out of the two is present in the final question vector. Hence, a question like `What does this teddy bear have on its neck?' won't signal the \emph{bear} category, while, `Does the bear love you?' will signal the \emph{bear} category. For this example, signalling the \emph{bear} category is an example of a false positive for the animal `bear' because the image actually has a `teddy bear' (see Figure \ref{fig:bears_false_positives}). In two word categories the order of the words is important. However, many questions have permutations of the two words. Owing to computational efficiency, these have not been included and the exact order which is present in the category type is used. Using an n-gram overlap for phrase word categories helps in differentiating cases like `hot dog' and  `dog'. \\

\begin{table*}[h]
\centering
\begin{tabular}{ccccccccccc}
\hline
What is the & What is  & Is the & Is this  & Is this a & Is there a 
& Is it & Is there  & Is & Is this an & Is that a  \\
\hline
\end{tabular}
\caption{Unconfirmed Question Types}
\label{unconfirmed_question_types}
\end{table*}

\begin{table*}[h]
\centering
\begin{tabular}{ccccc}
\hline
How many & What & What color is the & Are the & What kind of \\
What type of &
What are the &
Where is the &
Does the &
What color are the\\
Are these &
Are there &
Which &
What is the man&
Are \\
How &
Does this &
What is on the & 
What does the &
How many people are\\
What is in the &
What is this &
Do &
What are & 
Are they \\
What time &
What sport is &
Are there any &
What color is &
Why\\
Where are the &
What color &
Who is &
What animal is & 
Do you \\
How many people are in &
What room is &
Has &
What is the woman &
Can you\\
Why is the &
What is the color of the &
What is the person &
Could &
Was \\
What number is &
What is the name &
What brand &
Is the person &
Is he\\
Is the man &
Is the woman &
Is this person  \\
\hline
\end{tabular}
\caption{Confirmed Question Types}
\label{question_types}
\end{table*}

\begin{figure*}[h]
\begin{tabular}{ccccc}
\includegraphics[height=60pt]{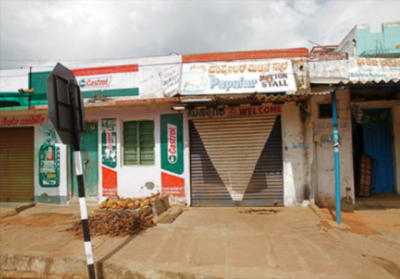} &
\includegraphics[height=60pt]{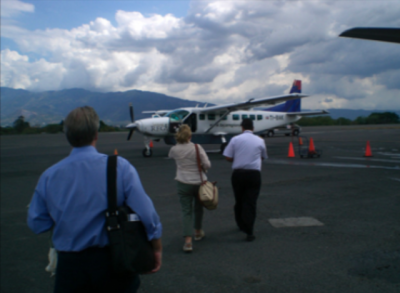} &
\includegraphics[height=60pt]{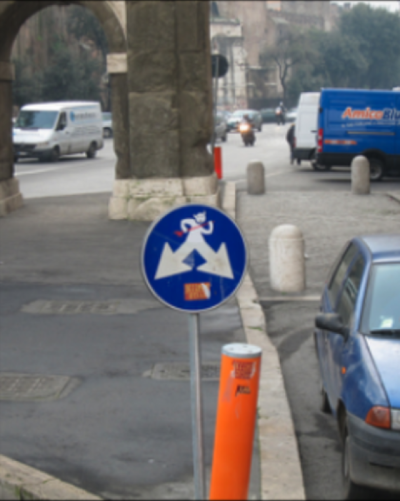} &
\includegraphics[height=60pt]{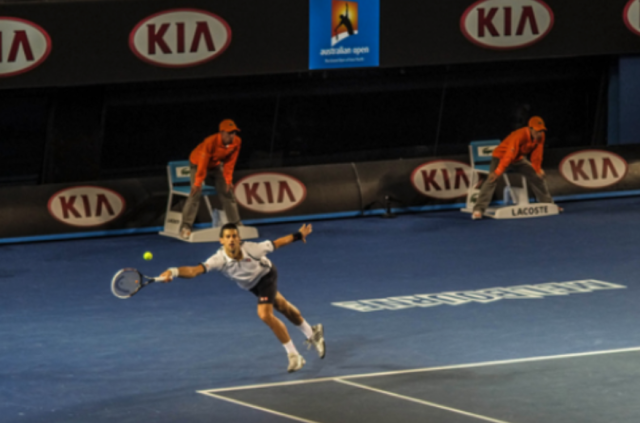} &
\includegraphics[height=60pt]{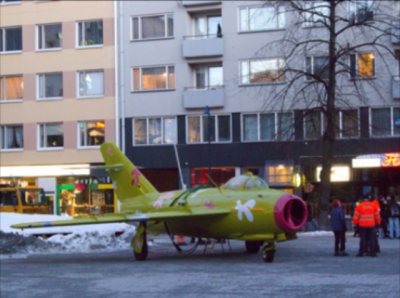} \\
\begin{minipage}[t]{0.20\linewidth}
\scriptsize \centering
What oil brand is on the building that is white, orange and green?
\end{minipage} &
\begin{minipage}[t]{0.17\linewidth}
\scriptsize \centering
How many orange cones \\ are there? 
\end{minipage} &
\begin{minipage}[t]{0.17\linewidth}
\scriptsize \centering
How many orange \\ poles are there?
\end{minipage} &
\begin{minipage}[t]{0.17\linewidth}
\scriptsize \centering
 What do the people \\ in orange do?
\end{minipage} &
\begin{minipage}[t]{0.17\linewidth}
\scriptsize \centering
How many people are wearing orange jackets? 
\end{minipage} \\
\end{tabular}
\caption{False Positives for the \emph{orange} category. As all the cases include `orange' as an adjective, these can be included in the algorithm by checking for POS tags.  }
\label{fig:orange_false_positives}
\end{figure*}

\smallsec{Ambiguous Words}
These include words with distinct meanings depending on the context. For instance, the category, \emph{orange}, is often mistaken with the color orange. There are more questions which use `orange' as an attribute rather than an object. Examples of false positives for \emph{orange} category are shown in Figure \ref{fig:orange_false_positives}, where `orange' is used as an adjective. \emph{Remote} is another such category, which is used as an adjective and as the target object. The animal category \emph{bear} is sometimes confused with the food item `gummy bears', and the sports team `Chicago Bears', due to fuzzy human annotated questions. The question `What is the percentage of yellow gummy bears?' will signal the category \emph{bear} and is a false positive. More false positives are shown in Figure \ref{fig:bears_false_positives}. \\ 

\smallsec{Confusing Categories}
Categories that frequently occur along with other categories confusing. For example, `handbag' instances co-occur with the \emph{person} category, \emph{remote} co-occurs with the \emph{tv} category and the `wii' object. Similarly, the \emph{toaster} category has 49 instances which co-occur with `microwave', `bowl' and `counter'. `Fork' co-occurs with various food categories causing confusion. `Dining table' gets confused with other food related categories.  \\

\smallsec{Less Instances}
Categories like \emph{toaster} have few instances and are often not asked about, directly. The \emph{oven} category often co-occurs with the \emph{microwave} and \emph{toaster} category. Examples include, `Is there a vintage toaster oven in the photo?', and `Where is the microwave oven?'. `Refrigerator' is confused with various categories related to food. It also co-occurs with the word `magnet' in most of its occurrences through questions like, `Are there magnets on the fridge?'.

\centering

\centering

\centering

\centering

{\small
\bibliographystyle{ieee}
\bibliography{egpaper_for_review}

\begin{thebibliography}{10}\itemsep=-1pt

\bibitem{spice2016}
P.~Anderson, B.~Fernando, M.~Johnson, and S.~Gould.
\newblock {SPICE}: Semantic propositional image caption evaluation.
\newblock In {\em European Conference on Computer Vision (ECCV)}, 2016.

\bibitem{Andreas-Rohrbach-2}
J.~Andreas, M.~Rohrbach, T.~Darrell, and D.~Klein.
\newblock Learning to compose neural networks for question answering.
\newblock In {\em North American Chapter of the Association for Computational
  Linguistics (NAACL)}, 2016.

\bibitem{VQA}
S.~Antol, A.~Agrawal, J.~Lu, M.~Mitchell, D.~Batra, C.~L. Zitnick, and
  D.~Parikh.
\newblock {VQA}: Visual question answering.
\newblock In {\em International Conference on Computer Vision (ICCV)}, 2015.

\bibitem{GIS_Assisted_ECCV14}
S.~Ardeshir, A.~Roshan~Zamir, and M.~Shah.
\newblock {GIS}-assisted object detection and geospatial localization.
\newblock In {\em European Conference on Computer Vision ({ECCV})}, 2014.

\bibitem{BaICCV15}
J.~Ba, K.~Swersky, S.~Fidler, and R.~Salakhutdinov.
\newblock Predicting deep zero-shot convolutional neural networks using textual
  descriptions.
\newblock In {\em International Conference on Computer Vision (ICCV)}, 2015.

\bibitem{chen-wang}
K.~Chen, J.~Wang, L.~Chen, H.~Gao, W.~Xu, and R.~Nevatia.
\newblock {ABC-CNN:} an attention based convolutional neural network for visual
  question answering.
\newblock {\em CoRR}, abs/1511.05960, 2015.

\bibitem{Urtasun14}
L.-C. Chen, S.~Fidler, A.~L. Yuille, and R.~Urtasun.
\newblock Beat the mturkers: Automatic image labeling from weak {3D}
  supervision.
\newblock In {\em Computer Vision and Pattern Recognition (CVPR)}, 2014.

\bibitem{NEIL}
X.~Chen, A.~Shrivastava, and A.~Gupta.
\newblock {NEIL}: {E}xtracting {V}isual {K}nowledge from {W}eb {D}ata.
\newblock In {\em International Conference on Computer Vision (ICCV)}, 2013.

\bibitem{learning_rvr_imgcap}
X.~Chen and C.~L. Zitnick.
\newblock Learning a recurrent visual representation for image caption
  generation.
\newblock In {\em Computer Vision and Pattern Recognition (CVPR)}, 2015.

\bibitem{seq2seq}
K.~Cho, B.~Van~Merri{\"{e}}nboer, {\c C}.~G{\"{u}}l{\c c}ehre, D.~Bahdanau,
  F.~Bougares, H.~Schwenk, and Y.~Bengio.
\newblock Learning phrase representations using rnn encoder--decoder for
  statistical machine translation.
\newblock In {\em Empirical Methods in Natural Language Processing (EMNLP)},
  2014.

\bibitem{meteor}
M.~Denkowski and A.~Lavie.
\newblock Meteor universal: Language specific translation evaluation for any
  target language.
\newblock In {\em EACL 2014 Workshop on Statistical Machine Translation}, 2014.

\bibitem{LEVAN}
S.~Divvala, A.~Farhadi, and C.~Guestrin.
\newblock Learning everything about anything: Webly-supervised visual concept
  learning.
\newblock In {\em Computer Vision and Pattern Recognition (CVPR)}, 2014.

\bibitem{lrcn2014}
J.~Donahue, L.~A. Hendricks, S.~Guadarrama, M.~Rohrbach, S.~Venugopalan,
  K.~Saenko, and T.~Darrell.
\newblock Long-term recurrent convolutional networks for visual recognition and
  description.
\newblock In {\em Computer Vision and Pattern Recognition (CVPR)}, 2015.

\bibitem{PASCALIJCV}
M.~Everingham, L.~Van~Gool, C.~K.~I. Williams, J.~Winn, and A.~Zisserman.
\newblock The {P}ascal {V}isual {O}bject {C}lasses ({VOC}) challenge.
\newblock {\em International Journal of Computer Vision (IJCV)},
  88(2):303--338, June 2010.

\bibitem{fang15}
H.~Fang, S.~Gupta, F.~Iandola, R.~K. Srivastava, L.~Deng, P.~Dollár, J.~Gao,
  X.~He, M.~Mitchell, J.~C. Platt, C.~L. Zitnick, and G.~Zweig.
\newblock From captions to visual concepts and back.
\newblock In {\em Computer Vision and Pattern Recognition (CVPR)}, 2015.

\bibitem{Fukui-park}
A.~Fukui, D.~H. Park, D.~Yang, A.~Rohrbach, T.~Darrell, and M.~Rohrbach.
\newblock Multimodal compact bilinear pooling for visual question answering and
  visual grounding.
\newblock In {\em Empirical Methods in Natural Language Processing (EMNLP)},
  2016.

\bibitem{GoyalMPB16}
Y.~Goyal, A.~Mohapatra, D.~Parikh, and D.~Batra.
\newblock Interpreting visual question answering models.
\newblock In {\em ICML Workshop on Visualization for Deep Learning}, 2016.

\bibitem{Andreas-Rohrbach}
L.~A. Hendricks, S.~Venugopalan, M.~Rohrbach, R.~Mooney, K.~Saenko, and
  T.~Darrell.
\newblock Deep compositional captioning: Describing novel object categories
  without paired training data.
\newblock In {\em Computer Vision and Pattern Recognition (CVPR)}, 2016.

\bibitem{nlpsegm16}
R.~Hu, M.~Rohrbach, and T.~Darrell.
\newblock Segmentation from natural language expressions.
\newblock In {\em European Conference on Computer Vision (ECCV)}, 2016.

\bibitem{Ilievski-yan}
I.~Ilievski, S.~Yan, and J.~Feng.
\newblock A focused dynamic attention model for visual question answering.
\newblock {\em CoRR}, abs/1604.01485, 2016.

\bibitem{JabriJM16}
A.~Jabri, A.~Joulin, and L.~van~der Maaten.
\newblock Revisiting visual question answering baselines.
\newblock In {\em European Conference on Computer Vision (ECCV)}, 2016.

\bibitem{jiang-wang}
A.~Jiang, F.~Wang, F.~Porikli, and Y.~Li.
\newblock Compositional memory for visual question answering.
\newblock {\em CoRR}, abs/1511.05676, 2015.

\bibitem{densecap}
J.~Johnson, A.~Karpathy, and L.~Fei-Fei.
\newblock Densecap: Fully convolutional localization networks for dense
  captioning.
\newblock In {\em Computer Vision and Pattern Recognition (CVPR)}, 2016.

\bibitem{Kafle_2016_CVPR}
K.~Kafle and C.~Kanan.
\newblock Answer-type prediction for visual question answering.
\newblock In {\em Computer Vision and Pattern Recognition (CVPR)}, 2016.

\bibitem{neuraltalk2}
A.~Karpathy and L.~Fei{-}Fei.
\newblock Deep visual-semantic alignments for generating image descriptions.
\newblock In {\em Computer Vision and Pattern Recognition (CVPR)}, 2015.

\bibitem{kim-lee}
J.~Kim, S.~Lee, D.~Kwak, M.~Heo, J.~Kim, J.~Ha, and B.~Zhang.
\newblock Multimodal residual learning for visual {QA}.
\newblock In {\em Computer Vision and Pattern Recognition (CVPR)}, 2016.

\bibitem{Krizhevsky12}
A.~Krizhevsky, I.~Sutskever, and G.~E. Hinton.
\newblock Imagenet classification with deep convolutional neural networks.
\newblock In F.~Pereira, C.~J.~C. Burges, L.~Bottou, and K.~Q. Weinberger,
  editors, {\em Advances in Neural Information Processing Systems (NIPS) 25},
  pages 1097--1105. Curran Associates, Inc., 2012.

\bibitem{COCO}
T.-Y. Lin, M.~Maire, S.~Belongie, J.~Hays, P.~Perona, D.~Ramanan, P.~Dollár,
  and C.~L. Zitnick.
\newblock {Microsoft COCO: Common Objects in Context}.
\newblock In {\em {European Conference on Computer Vision (ECCV)}}, 2014.

\bibitem{Lin_2016_ECCV}
X.~Lin and D.~Parikh.
\newblock Leveraging visual question answering for image-caption ranking.
\newblock In {\em European Conference on Computer Vision (ECCV)}, 2016.

\bibitem{nltk}
E.~Loper and S.~Bird.
\newblock Nltk: The natural language toolkit.
\newblock In {\em Workshop on Effective Tools and Methodologies for Teaching
  Natural Language Processing and Computational Linguistics}, 2002.

\bibitem{lu-yang}
J.~Lu, J.~Yang, D.~Batra, and D.~Parikh.
\newblock Hierarchical question-image co-attention for visual question
  answering.
\newblock In {\em Computer Vision and Pattern Recognition (CVPR)}, 2016.

\bibitem{Misra15}
I.~Misra, C.~L. Zitnick, M.~Mitchell, and R.~Girshick.
\newblock {Seeing through the Human Reporting Bias: Visual Classifiers from
  Noisy Human-Centric Labels}.
\newblock In {\em Computer Vision and Pattern Recognition (CVPR)}, 2016.

\bibitem{Noh-han}
H.~Noh and B.~Han.
\newblock Training recurrent answering units with joint loss minimization for
  {VQA}.
\newblock In {\em Computer Vision and Pattern Recognition (CVPR)}, 2016.

\bibitem{owens16}
A.~Owens, J.~Wu, J.~H. McDermott, W.~T. Freeman, and A.~Torralba.
\newblock Ambient sound provides supervision for visual learning.
\newblock In {\em European Conference on Computer Vision (ECCV)}, 2016.

\bibitem{Prest12}
A.~Prest, C.~Leistner, J.~Civera, C.~Schmid, and V.~Ferrari.
\newblock Learning object class detectors from weakly annotated video.
\newblock In {\em Computer Vision and Pattern Recognition (CVPR)}, 2012.

\bibitem{ILSVRC}
O.~Russakovsky, J.~Deng, H.~Su, J.~Krause, S.~Satheesh, S.~Ma, Z.~Huang,
  A.~Karpathy, A.~Khosla, M.~Bernstein, A.~C. Berg, and L.~Fei-Fei.
\newblock {ImageNet Large Scale Visual Recognition Challenge}.
\newblock {\em International Journal of Computer Vision (IJCV)}, 2015.

\bibitem{Saito-shin}
K.~Saito, A.~Shin, Y.~Ushiku, and T.~Harada.
\newblock Dualnet: Domain-invariant network for visual question answering.
\newblock {\em CoRR}, abs/1606.06108, 2016.

\bibitem{Shih-singh}
K.~J. Shih, S.~Singh, and D.~Hoiem.
\newblock Where to look: Focus regions for visual question answering.
\newblock In {\em Computer Vision and Pattern Recognition (CVPR)}, 2016.

\bibitem{Szegedy14}
C.~Szegedy, W.~Liu, Y.~Jia, P.~Sermanet, S.~Reed, D.~Anguelov, D.~Erhan, and
  A.~Rabinovich.
\newblock Going deeper with convolutions.
\newblock In {\em Computer Vision and Pattern Recognition (CVPR)}, 2015.

\bibitem{Vendrov16}
I.~Vendrov, R.~Kiros, S.~Fidler, and R.~Urtasun.
\newblock Order-embeddings of images and language.
\newblock In {\em International Conference on Learning Representations (ICLR)},
  2016.

\bibitem{Vinyals14}
O.~Vinyals, A.~Toshev, S.~Bengio, and D.~Erhan.
\newblock Show and tell: A neural image caption generator.
\newblock In {\em Computer Vision and Pattern Recognition (CVPR)}, 2015.

\bibitem{weinzaepfel2016towards}
P.~Weinzaepfel, X.~Martin, and C.~Schmid.
\newblock Towards weakly-supervised action localization.
\newblock {\em arXiv preprint arXiv:1605.05197}, 2016.

\bibitem{wu-wang}
Q.~Wu, P.~Wang, C.~Shen, A.~Dick, and A.~{van den Hengel}.
\newblock Ask me anything: free-form visual question answering based on
  knowledge from external sources.
\newblock In {\em Computer Vision and Pattern Recognition (CVPR)}, 2016.

\bibitem{Xiong-merity}
C.~Xiong, S.~Merity, and R.~Socher.
\newblock Dynamic memory networks for visual and textual question answering.
\newblock In {\em International Conference on Machine Learning (ICML)}, 2016.

\bibitem{xu-saenko}
H.~Xu and K.~Saenko.
\newblock Ask, attend and answer: Exploring question-guided spatial attention
  for visual question answering.
\newblock In {\em European Conference on Computer Vision (ECCV)}, 2016.

\bibitem{Xu15}
J.~Xu, A.~G. Schwing, and R.~Urtasun.
\newblock Learning to segment under various forms of weak supervision.
\newblock In {\em Computer Vision and Pattern Recognition (CVPR)}, 2015.

\bibitem{yang-he}
Z.~Yang, X.~He, J.~Gao, L.~Deng, and A.~Smola.
\newblock Stacked attention networks for image question answering.
\newblock In {\em Computer Vision and Pattern Recognition (CVPR)}, 2016.

\bibitem{bow-bolei}
B.~Zhou, Y.~Tian, S.~Sukhbaatar, A.~Szlam, and R.~Fergus.
\newblock Simple baseline for visual question answering.
\newblock {\em CoRR}, abs/1512.02167, 2015.

\bibitem{ZhuICCV15}
Y.~Zhu, R.~Kiros, R.~Zemel, R.~Salakhutdinov, R.~Urtasun, A.~Torralba, and
  S.~Fidler.
\newblock Aligning books and movies: Towards story-like visual explanations by
  watching movies and reading books.
\newblock In {\em International Conference on Computer Vision (ICCV)}, 2015.

\end{thebibliography}
}

\end{document}